\documentclass[runningheads]{llncs}
\usepackage{xcolor}
\usepackage{amsmath}
\usepackage[T1]{fontenc}
\usepackage{graphicx}
\usepackage{listings}
\usepackage{xspace}
\usepackage{booktabs}
\usepackage{multirow}
\usepackage{subcaption}

\usepackage{caption}

\usepackage{float} 

\newcommand{\mypara}[1]{\vspace{2pt}\noindent{\textit{\textbf{#1}}}\xspace}

\usepackage{xcolor}

\lstdefinestyle{customc}{
  language=C,
  numbers=left,                     
  numberstyle=\tiny\color{gray},   
  stepnumber=1,                    
  numbersep=10pt,                  
  backgroundcolor=\color{white},  
  showspaces=false,                
  showstringspaces=false,          
  showtabs=false,                  
  frame=false,    
  rulecolor=\color{black},         
  tabsize=2,                       
  captionpos=b,                    
  breaklines=true,                 
  breakatwhitespace=true,          
  keywordstyle=\color{blue},       
  commentstyle=\color{gray}\itshape, 
  stringstyle=\color{orange},      
  basicstyle=\ttfamily\small,      
}

\begin{document}
\title{Decentor-V: Lightweight ML Training on Low-Power RISC-V Edge Devices}

%
\author{Marcelo Ribeiro \and
Diogo Costa \and
Gonçalo Moreira \and
Sandro Pinto \and
Tiago Gomes}

\authorrunning{M. Ribeiro et al.}

\institute{Centro ALGORITMI / LASI, Universidade do Minho, Portugal \\
\email{\{pg54028, id10560, pg53841\}@alunos.uminho.pt, \
\{sandro.pinto, mr.gomes\}@dei.uminho.pt}}

\maketitle              
\begin{abstract}

Modern IoT devices increasingly rely on machine learning solutions to process data locally. However, the lack of graphics processing units (GPUs) or dedicated accelerators on most platforms makes on-device training largely infeasible, often requiring cloud-based services to perform this task. This procedure often raises privacy-related concerns, and creates dependency on reliable and always-on connectivity. 
Federated Learning (FL) is a new trend that addresses these issues by enabling decentralized and collaborative training directly on devices, but it requires highly efficient optimization algorithms. 
L-SGD, a lightweight variant of stochastic gradient descent, has enabled neural network training on Arm Cortex-M Microcontroller Units (MCUs). 
{This work extends L-SGD to RISC-V-based MCUs, an open and emerging architecture that still lacks robust support for on-device training. L-SGD was evaluated on both Arm and RISC-V platforms using 32-bit floating-point arithmetic, highlighting the performance impact of the absence of Floating-Point Units (FPUs) in RISC-V MCUs. To mitigate these limitations, we introduce an 8-bit quantized version of L-SGD for RISC-V, which achieves nearly $4\times$ reduction in memory usage and a $2.2\times$ speedup in training time, with negligible accuracy degradation.}

\end{abstract}
%
%
\section{Introduction}

The rapid expansion of the Internet of Things (IoT) is reshaping the digital landscape, with billions of interconnected devices generating and processing data in real time~\cite{IOT}. These devices are being increasingly deployed in domains such as smart cities~\cite{CITIES}, precision agriculture~\cite{AGRIC}, industrial automation~\cite{INDUSTRY}, and personalized healthcare~\cite{HEALTH}, where responsiveness, data privacy, and reliability are critical~\cite{EDGECOMPUTING}. However, traditional machine learning (ML) approaches typically depend on centralized cloud infrastructure for model training and inference, requiring the transmission of large volumes of raw data over the network~\cite{EDGECOMPUTING}. This model not only introduces communication bottlenecks and latency but also raises significant privacy and regulatory concerns, particularly under legal regulations such as the General Data Protection Regulation (GDPR)~\cite{GDPR}.

To address these limitations, a shift toward decentralized intelligence has emerged~\cite{trainmeifyoucan}. Edge-based training allows models to be updated locally on the device, preserving data locality, reducing dependency on persistent connectivity, and minimizing exposure of sensitive information~\cite{local}. Federated Learning (FL) embodies this paradigm~\cite{federated_learning_survey}, enabling collaborative model updates across a network of devices without sharing raw data~\cite{FL1}.
Nonetheless, enabling on-device learning in highly constrained environments, where memory, processing power, and energy are limited, remains an open challenge. Traditional optimization algorithms~\cite{gd,SGD,adagrad,adam} are not suitable for MCUs~\cite{lw}, which typically may lack floating-point units (FPUs) and offer only a fraction of the resources available on desktop or cloud platforms. To bridge this gap, L-SGD~\cite{decentor}, a lightweight variant of stochastic gradient descent (SGD)~\cite{SGD}, was previously introduced for the Arm Cortex-M family, demonstrating efficient training under severe resource constraints~\cite{trainmeifyoucan}.

This article extends L-SGD to RISC-V-based MCUs, an open and increasingly adopted instruction-set architecture (ISA) in the embedded domain~\cite{riscv_survey}. 
Its modular design and growing ecosystem make it a strong candidate for future edge devices, yet it still lacks optimized support for on-device ML training. 
This work explores the feasibility of quantized local training on RISC-V-based platforms, analyzing its performance under realistic memory and computational constraints, and comparing it against existing Arm-based implementations. 
The key contributions of this work are as follows:

\begin{itemize}
    \item \textbf{Portability of L-SGD to RISC-V}: To the best of the authors' knowledge, this is the first suitable implementation of local ML training on RISC-V-based MCUs.

    \item \textbf{Cross-platform performance analysis}: A comparative evaluation of L-SGD is conducted across Arm Cortex-M and RISC-V MCUs.
    
    \item \textbf{Quantization impact}: Results show that 8-bit quantization leads up to near $4\times$ reduction of memory footprint and up to a $2.2\times$ improvement in execution time on RISC-V with negligible performance degradation in terms of precision, recall, F1-Score, and accuracy.
\end{itemize}

\section{Background and Related Work}

\subsection{Machine Learning Training}

Training a ML model involves teaching it to make accurate predictions by minimizing the error between its outputs and the expected results. In supervised learning~\cite{supervised}, this process begins with a dataset composed of input samples, each paired with a corresponding label, which represents the correct output the model should learn to predict. When an input is fed into the model, it produces an output based on its current internal parameters, known as weights and bias. This output is then compared to the true label, and the difference between them, called the error, is used to calculate the loss, a numerical value that quantifies how far off the model's prediction is.
The goal of training is to reduce this loss. The smaller the loss, the more accurate the model's predictions are. To achieve this, optimization algorithms, commonly referred to as optimizers, are used. These optimizers adjust the model’s weights and bias in a way that incrementally reduces the loss, guiding the model towards better performance over time. This process is repeated over many iterations, gradually improving the model's ability to learn from the training data to new, unseen inputs.

The most fundamental of these optimizers is Gradient Descent (GD) \cite{gd}, which computes the gradient of the entire dataset to guide parameter updates. While effective in theory, GD becomes computationally expensive and impractical for large-scale or streaming datasets. To mitigate this issue, the Stochastic Gradient Descent (SGD) optimizer \cite{SGD} was introduced, which updates weights and bias using individual or small mini-batches of samples. This reduces the computational cost per update step and enables better scalability, particularly for more demanding training scenarios.
Building on SGD, more advanced optimizers like Adagrad \cite{adagrad} and Adam \cite{adam} have been developed. These methods adapt the learning rate dynamically for each parameter based on the history of gradients, often leading to faster convergence and better performance, especially on complex or noisy tasks. However, these adaptive optimizers require additional memory to store auxiliary variables, such as gradient squares or momentum terms, which increases both memory footprint and computational overhead during training.

These requirements pose significant challenges when training models on typical resource-constrained devices such as MCUs, which typically lack FPUs, have very limited Random Access Memory (RAM) and flash memory, and must operate under tight energy budgets. To address these limitations, Lightweight Stochastic Gradient Descent (L-SGD)~\cite{trainmeifyoucan} was proposed as an optimized variant of SGD specifically designed for on-device training on MCUs. L-SGD introduces a technique called node delta optimization, which aggregates errors at the neuron level rather than per connection. This dramatically reduces memory usage and simplifies the back-propagation process, enabling real-time, low-power training without relying on cloud infrastructure. Consequently, L-SGD opens the door for practical, local learning on edge devices, which is critical for applications requiring privacy, low latency, or offline operation.

\subsection{Related Work}

\mypara{Deep Edge ML Inference:}  
Several software libraries have been developed to support deep learning inference on resource-constrained devices. CMSIS-NN~\cite{CMSIS-NN}, developed by Arm, provides efficient neural network kernels optimized for Cortex-M processors. Similarly, PULP-NN~\cite{pulp_nn}, a library built for the Parallel Ultra-Low Power (PULP) platform~\cite{pulp_platform}, offers hand-optimized kernels for quantized deep neural network (DNN) inference on ultra-low-power RISC-V cores.
Also, an inference-only extension for quantized Capsule Networks on MCUs has been proposed on both Arm Cortex-M and RISC-V MCUs~\cite{capsnet_edge}. This work includes custom software kernels and a post-training quantization framework, enabling {\textit{int8}} CapsNet inference with significant memory savings and low latency on MCU-based platforms.

\mypara{Deep Edge ML Training:}  
Training ML models directly on embedded devices has gained attention as a way to improve data privacy, reduce communication costs, and enable adaptive systems. Costa et al.~\cite{trainmeifyoucan} introduced L-SGD, a lightweight variant of SGD tailored for edge devices with optimizations for memory and computation. {In addition, their work included a comparative study of L-SGD against several widely used optimizers (e.g., GD, SGD, Adam, and Adagrad), demonstrating its suitability for resource-constrained environments.} Building on this, Decentor~\cite{decentor} demonstrated that decentralized training is feasible on ultra-low-power Arm Cortex-M MCU, showing that even constrained devices like the Cortex-M0+ can support local training of compact neural networks. These efforts illustrate the growing potential for edge-based training in real-world applications.

\mypara{Gap Analysis:}  
While most prior work focuses on either inference optimization (e.g., CMSIS-NN, PULP-NN) or Arm-based on-device training \cite{decentor}, there remains a gap in platform-independent training support across emerging architectures like RISC-V. 
The gap is addressed by adapting L-SGD for portable deployment on generic RISC-V-based MCUs. 
In contrast to prior solutions tied to specific hardware features or vendors, this implementation emphasizes modularity and architectural abstraction. This enables support for quantized training workflows even on devices lacking FPUs or advanced memory hierarchies, broadening the applicability of deep edge training methods on RISC-V. 

\section{L-SGD in RISC-V}

To support the execution of L-SGD on RISC-V-based MCUs, this work introduces a lightweight and modular implementation that leverages the core features of the RISC-V ISA~\cite{riscv_isa}.
The design is independent of any specific hardware platform, relying solely on standard RISC-V instructions. 
The core components of the implementation include a custom fixed-point arithmetic module, an optimized loop structure for gradient updates, and a lightweight memory management strategy to store model parameters and intermediate values.
The following subsections describe the key modifications introduced to adapt the original ARM-based L-SGD implementation for RISC-V.

\subsection{System Overview}

\begin{figure}[t]
    \centering
    \begin{subfigure}[b]{0.49\textwidth}
        \centering
        \includegraphics[width=\textwidth]{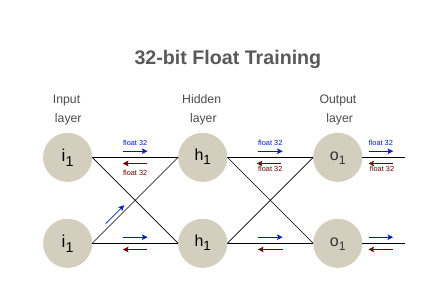}
        \caption{Float32 Training workflow.}
        \label{fig:TrainingWorkflowA}
    \end{subfigure}
    \hfill
    \begin{subfigure}[b]{0.49\textwidth}
        \centering
        \includegraphics[width=\textwidth]{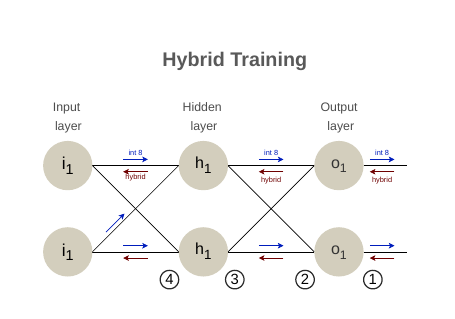}
        \caption{Hybrid Training workflow.}
        \label{fig:TrainingWorkflowB}
    \end{subfigure}
    \caption{{Training workflow.}}
    \label{fig:Training Workflow}
\end{figure}

{To bridge the gap between performance and accuracy, this work proposes a hybrid quantized training strategy, as illustrated in Figure~\ref{fig:Training Workflow}. Figure~\ref{fig:TrainingWorkflowA} depicts the conventional {\textit{float32}} training pipeline, while Figure~\ref{fig:TrainingWorkflowB} shows the proposed hybrid approach. In this hybrid method, the forward pass is performed entirely in {\textit{int8}}, including quantized weights, biases, activation functions, and matrix operations, thereby providing high efficiency. 
However, to overcome the limitations of pure {\textit{int8}} training, the backward pass selectively reverts to {\textit{float32}} arithmetic: \textcircled{1} the quantized output is dequantized and the loss is computed in {\textit{float32}} to avoid saturation errors; \textcircled{2} the weight and bias updates are calculated and applied in {\textit{float32}}; \textcircled{3} the backpropagated deltas are computed using {\textit{float32}} to prevent the saturation and precision issues of {\textit{int8}}; and \textcircled{4} the activations for backward propagation are passed through {\textit{float32}} partial derivatives of the activation functions.}

Although the hybrid approach introduces additional quantization and dequantization steps, it mitigates the performance penalty of implementing L-SGD on platforms without FPUs by shifting all operations that can be safely quantized to {\textit{int8}}, while retaining {\textit{float32}} only where necessary to preserve training stability. 
This reduces the memory footprint to meet the constraints of resource-limited devices, providing a balanced solution that preserves accuracy and stability during training, ensuring efficiency and deployability on embedded hardware.

\subsection{L-SGD port to RISC-V}\label{sec:design}

\subsection*{Replacement of Standard Math Functions}

Standard math library functions such as \textit{exp()} and \textit{pow()}, commonly used in ML inference and training, are computationally expensive and rely on floating-point operations. However, RISC-V MCUs, like many other low-power platforms, often lack an FPU, resulting in software emulation of these operations and incurring significant performance and energy overheads.
To overcome this limitation, we developed lightweight, FPU-independent alternatives optimized for low-power MCUs. 
Specifically, we introduce \textit{fast\_round()}, \textit{fast\_power\_of\_two()}, and \textit{fast\_exp()}, efficient approximations of the standard \textit{round()}, \textit{pow()}, and \textit{exp()} functions, respectively, designed to maintain accuracy while significantly reducing computational cost.

\mypara{\textit{fast\_round()}:}Rounds a float to the nearest integer using simple arithmetic and casting, avoiding slower library functions.

\mypara{\textit{fast\_power\_of\_two()}:} Returns powers of two using left shift operations, which are computationally efficient on integer hardware.

\mypara{\textit{fast\_exp()}:} Approximates the exponential function using type punning and a precomputed scaling factor. It scales the input and reinterprets the result as a float using IEEE 754 encoding, avoiding expensive math operations. While slightly less accurate than standard methods, this technique greatly improves speed and is suitable for on-device training, where efficiency matters most. 

\subsection*{Added code on \textit{pulp\_nn\_linear\_int8()} for Single-Neuron Cases}\label{subsec:pulp_function}
In the paper version of the PULP-NN library used in this work, the function used to implement fully connected layers -- \textit{pulp\_nn\_linear\_int8()} -- was modified to properly support the case where the number of output neurons is one. The original implementation assumed multiple outputs and failed to handle this scenario correctly, which is common in compact models or binary classification tasks. The modified version initializes the accumulation with the bias (shifted and rounded), computes the dot product in a single loop, applies quantization scaling, clips the result to 8-bit range, and writes it to the output. This fix ensures correct behavior for single-neuron outputs in fully connected layers.

\subsection{Quantized training}\label{sec:quantizationimple}

To enable efficient training on RISC-V architectures, the DNN contains all its model parameters, including weights and biases, in quantized format. The forward pass is entirely quantized using fixed-point arithmetic, leveraging the fixed-point fully connected layers introduced previously in \ref{subsec:pulp_function}. For non-linearities, we use look-up tables (LUTs) to approximate activation functions at each layer, allowing for fast and memory-efficient evaluation.
Training a DNN from scratch using quantized representations on embedded hardware poses significant challenges, primarily due to error saturation during the early stages of training~\cite{saturat}. At the start of training, when accuracy is close to 0\%, the model's predictions are usually far from the correct labels, leading to very high loss values. These large loss values, when backpropagated through the quantized network, must be dequantized to fit a fixed range (e.g., \([-1, 1]\)). However, since these values are too large, they exceed this range and are clipped (saturated) at the boundaries, causing a loss of resolution. This saturation propagates backward through the network, causing both loss of numerical resolution and the potential introduction of severe quantization errors, such as overflows that may invert the sign of gradients. Together, these effects distort the learning signal and can critically impair the training process.

To mitigate this issue, all quantized models used must be initialized with pre-trained weights and bias. This strategy avoids the unstable and highly sensitive early training phase, thereby reducing both convergence time and error magnitude. 
Instead of training the model entirely from scratch, fine-tuning is applied to the quantized network. 
This represents a well-suited scenario for practical deployment in FL, where model personalization and on-device updates are prioritized over full retraining~\cite{personFL}.
A further challenge arises due to saturation effects when using {\textit{int8}} arithmetic for activation functions, their partial derivatives, and loss functions. 
These components often involve values that naturally extend beyond a fixed-point range (e.g., [-1, 1]), but in quantized implementations, such values are clipped at the boundaries, leading to a loss of gradient resolution and a large quantization error. 
To address this, a hybrid solution is implemented in the backward pass. Instead of storing a fully dequantized version of the model in memory, which would significantly increase resource consumption, a selective dequantization strategy is applied, processing one layer at a time during activation functions, their partial derivatives, and gradient computations. 
This approach strikes a balance between maintaining sufficient numerical precision for effective learning and minimizing memory usage, thereby meeting the tight hardware constraints of embedded platforms.

\section{Evaluation}\label{sec:evaluation}

The evaluation assesses the training and inference capabilities of the L-SGD algorithm on resource-constrained MCUs, focusing on both floating-point and quantized scenarios. The goal is to analyze the trade-offs between training quality, performance, and execution time, across different hardware platforms and numeric representations.

\mypara{Methodology:}
 The analysis was conducted using two implementations of the L-SGD algorithm: the \textit{default\_l-sgd} version, evaluated on both Arm and RISC-V platforms, and the \textit{opt\_l-sgd} version evaluated exclusively on RISC-V, as the optimizations specifically target architectural constraints and performance bottlenecks unique to RISC-V. 
 To ensure fair and consistent comparisons across implementations, data shuffling was disabled during training. 

\mypara{Datasets:}
{For the evaluation, two datasets were used: (i) CogDist \cite{cogdist_dataset}, and the (ii) CarEvaluation \cite{car_dataset}, whose models are detailed in Table \ref{tab:ml_models}.
CogDist is a binary classification task aimed at detecting cognitive distraction using sensor data. It consists of 3600 data points, each characterized by 6 input features representing sensor measurements.
In contrast, CarEvaluation is a multi-class classification task evaluates car acceptability based on categorical attributes. This dataset comprises 1728 data points, also described by 6 input features.
These datasets were chosen as good representatives of low-complexity models that are suitable to run on sensors based on resource-constrained MCUs.} 

\begin{table}[t]
\centering
\caption{Machine Learning Models (FC stands for Fully-Connected layers).}
\label{tab:ml_models}
\begin{tabular}{l c c c c c c c}
\toprule
\textbf{Dataset} & \textbf{Input} & \textbf{Layer 0} & \textbf{Act.} & \textbf{Layer 1} & \textbf{Act.} & \textbf{Layer 2} & \textbf{Act.} \\
\midrule
CogDist       & \shortstack{Input\\(6)}  & \shortstack{FC\\(40)} & TanH & \shortstack{FC\\(32)} & TanH & \shortstack{FC\\(1)}  & Sigmoid \\
\midrule
CarEvaluation & \shortstack{Input\\(6)}  & \shortstack{FC\\(32)} & TanH & \shortstack{FC\\(16)} & TanH & \shortstack{FC\\(4)}  & Sigmoid \\
\bottomrule
\end{tabular}
\end{table}

\mypara{Hardware Platforms and Toolchains:}
 All experiments were carried out on the STM32F767ZI~\cite{STM} (Arm Cortex-M7), which contains an FPU, and the GAP-8 V1~\cite{GAP8} (RISC-V), which lacks this hardware feature. Both platforms were configured to run on a single core at 216 MHz. 
 The code was compiled using the \textit{arm-none-eabi-gcc} and \textit{riscv32-unknown-elf-gcc} toolchains for the Arm and RISC-V targets, respectively.

\subsection{L-SGD in RISC-V (float-32)}

\begin{figure}[t]
    \includegraphics[width=\textwidth]{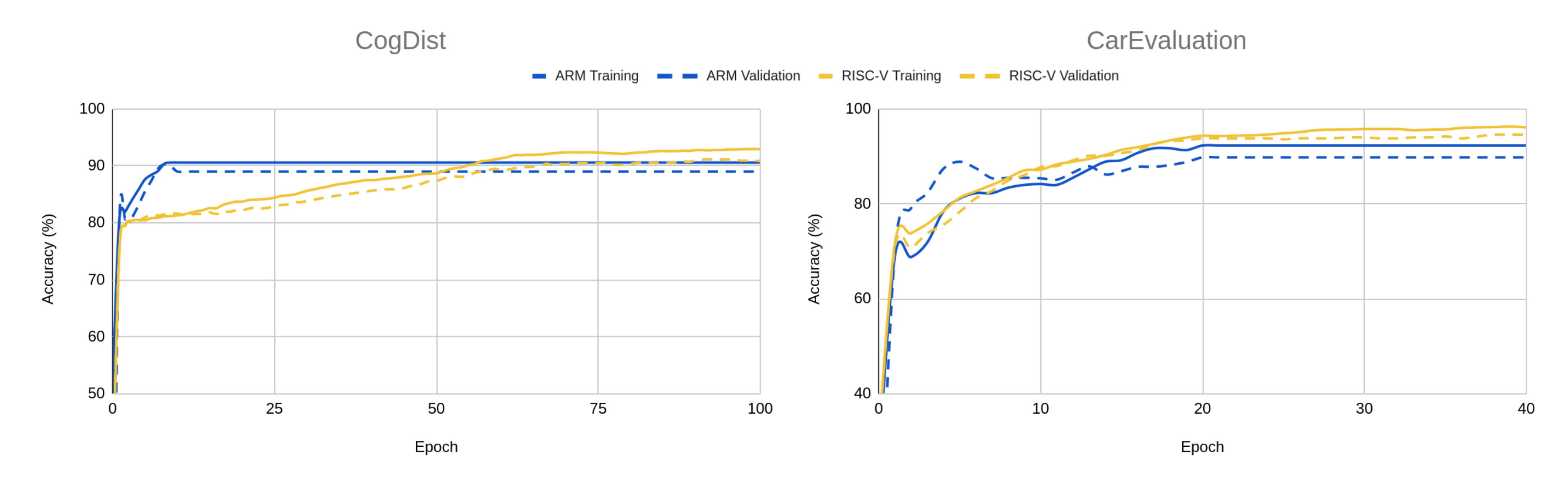}
    \caption{32-bit Float training curves.} 
    \label{fig:32-bit Float datasets training curves}
\end{figure}

\begin{table}[t]
\centering
\caption{{Float32 training performance on Arm and RISC-V architectures.}}
\renewcommand{\arraystretch}{1.2}
\resizebox{\linewidth}{!}{%
\begin{tabular}{l@{\hskip 0.5cm}l@{\hskip 0.5cm}c@{\hskip 0.5cm}c@{\hskip 0.5cm}c@{\hskip 0.5cm}c}
\hline
\textbf{Dataset} & \multicolumn{1}{c@{\hskip 0.3cm}}{\textbf{L-SGD Configuration}} & \textbf{Precision} & \textbf{Recall} & \textbf{F1-Score} & \textbf{Accuracy} \\
\hline
\multirow{3}{*}{CogDist} 
& (float32) Arm default\_l-sgd    & 89.00\% & 92.20\% & 90.60\% & 90.60\% \\ 
\cline{2-6}
& (float32) RISC-V default\_l-sgd & 91.75\% & 94.35\% & 93.03\% & 92.96\% \\ 
\cline{2-6}
& (float32) RISC-V opt\_l-sgd     & 91.75\% & 94.35\% & 93.03\% & 92.96\% \\ 
\hline
\multirow{3}{*}{CarEvaluation} 
& (float32) Arm default\_l-sgd    & 89.80\% & 81.70\% & 85.60\% & 93.10\% \\ 
\cline{2-6}
& (float32) RISC-V default\_l-sgd & 84.87\% & 94.45\% & 88.44\% & 96.10\% \\ 
\cline{2-6}
& (float32) RISC-V opt\_l-sgd     & 84.87\% & 94.45\% & 88.44\% & 96.10\% \\
\hline
\end{tabular}%
}
\label{tab:performance-float32}
\end{table}

\mypara{Training and Performance:}
Figure~\ref{fig:32-bit Float datasets training curves} illustrates the training and validation accuracy curves for the CogDist and CarEvaluation datasets, showing that both Arm and RISC-V platforms converge stably with closely aligned trajectories. In CogDist, both implementations converge to around 92\% accuracy, with RISC-V exhibiting a slightly more gradual progression. For CarEvaluation, both platforms surpass 95\% accuracy, with minimal divergence between training and validation curves.
Table~\ref{fig:32-bit Float datasets training curves} complements these results by summarizing the performance of the {\textit{float32}} implementation of L-SGD on the CogDist and CarEvaluation datasets for both Arm and FPU-less RISC-V platforms. On CogDist, Arm achieves about 91\% accuracy with an F1-score of 91\%, while RISC-V reaches 93\% accuracy and an F1-score of 93\%, showing slightly higher performance despite lacking an FPU. On CarEvaluation, RISC-V again outperforms Arm, achieving 96\% accuracy and an 88\% F1-score compared to 93\% and 86\% on Arm. These results confirm that the RISC-V implementation delivers training stability and learning behavior comparable to Arm, with only minor differences in convergence dynamics. The use of optimized mathematical functions on RISC-V preserves classification quality across all metrics, demonstrating that RISC-V MCUs can reliably support \textit{float32} training at a level on par with, and in some cases exceeding, Arm-based systems.

\mypara{Execution performance:}
Table~\ref{tab:exec_time_fpu} reports the average execution time per sample for {\textit{float32}} training on the CogDist and CarEvaluation datasets, comparing the Arm-based STM32F767ZI and the RISC-V-based GAP-8 V1 platforms. 
As expected, RISC-V shows higher latency than Arm with hardware FPU enabled, e.g., for CogDist, execution time rises from 4.8 ms to 17 ms (3.7×), and for CarEvaluation from 2.4 ms to 11 ms (4.6×). 
To isolate the effect of the FPU, the same workloads were executed on the Arm platform with the FPU disabled, resulting in 30 ms for CogDist and 19.6 ms for CarEvaluation, both worse than the GAP-8. 
This confirms that the main source of performance degradation on the GAP-8 is the absence of a hardware FPU rather than other architectural factors. Despite this limitation, the GAP-8 delivers reliable {\textit{float32}} training with acceptable execution time, demonstrating viability of RISC-V MCUs for on-device learning in scenarios where energy efficiency and flexibility are prioritized over raw speed.

\begin{table}[t]
\centering
\caption{{Average execution time per sample for float32 training.}}
\label{tab:exec_time_fpu}
\begin{tabular}{@{}l l c@{\hskip 0.6cm}c@{}}
\toprule
\textbf{Dataset} & \textbf{L-SGD Configuration} & \textbf{With FPU} & \textbf{Without FPU} \\
\midrule

\multirow{2}{*}{\textit{CogDist}} 
  & (float32) ARM \textit{default\_l-sgd}    & 4.79 ms  & 30 ms   \\
  & (float32) RISC-V \textit{opt\_l-sgd}     & ---     & 17 ms   \\

\midrule

\multirow{2}{*}{\textit{CarEvaluation}} 
  & (float32) ARM \textit{default\_l-sgd}    & 2.38 ms  & 19.61 ms \\
  & (float32) RISC-V \textit{opt\_l-sgd}     & ---     & 11 ms    \\

\bottomrule
\end{tabular}
\end{table}

\subsection{L-SGD in RISC-V (int-8)}

\begin{figure}[t]
    \includegraphics[width=\textwidth]{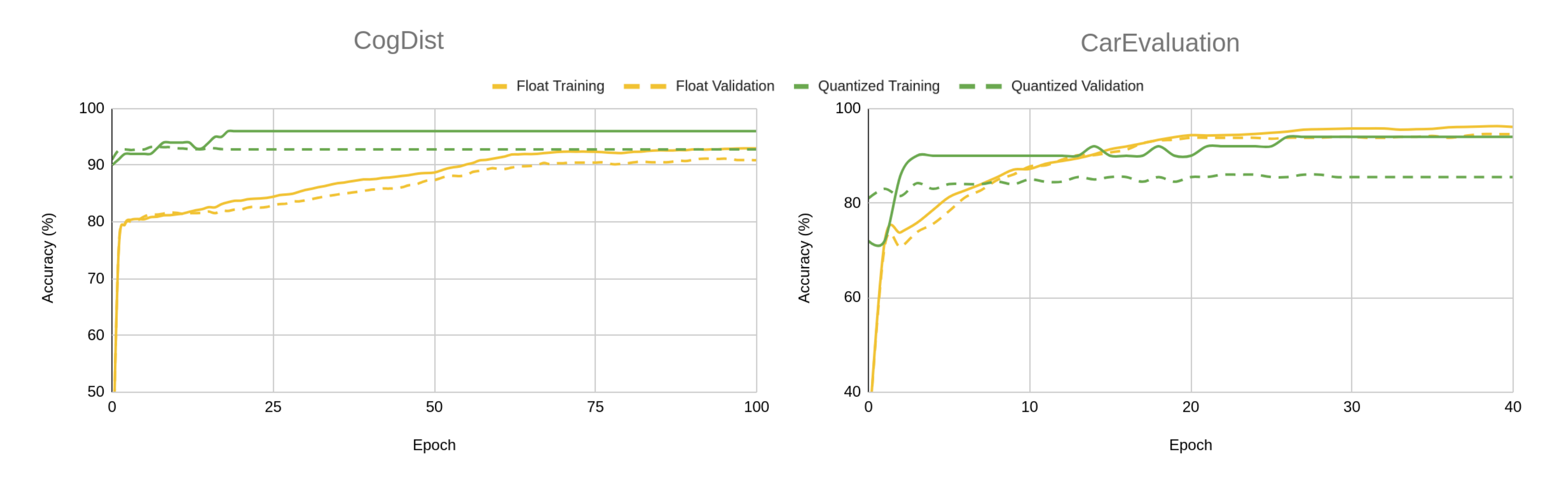}
    \caption{RISC-V training curves} 
    \label{fig:RISC-V training curves}
\end{figure}
   
\begin{table}[t]
\centering
\caption{{Performance results for float32 and int8 training on RISC-V.}}
\renewcommand{\arraystretch}{1.2}
\resizebox{\linewidth}{!}{%
\begin{tabular}{l@{\hskip 0.5cm}l@{\hskip 0.5cm}c@{\hskip 0.5cm}c@{\hskip 0.5cm}c@{\hskip 0.5cm}c} 
\hline
\textbf{Dataset} & \multicolumn{1}{c@{\hskip 0.3cm}}{\textbf{L-SGD Configuration}} & \textbf{Precision} & \textbf{Recall} & \textbf{F1-Score} & \textbf{Accuracy} \\ 
\hline
\multirow{3}{*}{CogDist} 
& (float32) RISC-V opt\_l-sgd   & 91.75\% & 94.35\% & 93.03\% & 92.96\% \\
\cline{2-6}
& (int8) RISC-V default\_l-sgd  & 94.00\% & 97.92\% & 95.92\% & 96.00\% \\
\cline{2-6}
& (int8) RISC-V opt\_l-sgd      & 94.00\% & 97.92\% & 95.92\% & 96.00\% \\
\hline
\multirow{3}{*}{CarEvaluation} 
& (float32) RISC-V opt\_l-sgd   & 84.87\% & 94.45\% & 88.44\% & 96.10\% \\
\cline{2-6}
& (int8) RISC-V default\_l-sgd  & 94.00\% & 94.00\% & 94.00\% & 94.00\% \\
\cline{2-6}
& (int8) RISC-V opt\_l-sgd      & 94.00\% & 94.00\% & 94.00\% & 94.00\% \\
\hline
\end{tabular}%
}
\label{tab:quantized_training_performance}
\end{table}

\mypara{Training and Performance:}

To evaluate the impact of quantization on training dynamics, Figure~\ref{fig:RISC-V training curves} shows the training and validation accuracy curves for the CogDist and CarEvaluation datasets, comparing {\textit{float32}} and {\textit{int8}} models on the RISC-V-based GAP-8 V1 platform. To mitigate saturation effects, both models were initialized from a pre-trained configuration. For CogDist (left plot), the {\textit{int8}} model rapidly surpasses 96\% training accuracy within 20 epochs, outperforming the {\textit{float32}} model, which converges near 92\% after nearly 70 epochs. Validation accuracy remains closely aligned, indicating promising generalization without overfitting. A similar trend appears in CarEvaluation (right plot), where the quantized model starts at 72\% accuracy and reaches 90\% about 10 epochs earlier than float. Its final accuracy reaches 94\%, closely matching the float model. These results show that {\textit{int8}} quantization maintains learning capability while accelerating early convergence. To further assess training performance, Table\ref{tab:quantized_training_performance} reports Precision, Recall, F1-Score, and Accuracy for both datasets. Quantized {\textit{int8}} training maintained performance comparable to {\textit{float32}} across both tasks: in the binary CogDist dataset, all configurations performed identically, while in the larger multi-class CarEvaluation task, quantized models again matched float accuracy, demonstrating effectiveness even under increased complexity.

\mypara{Execution Time:}
Table~\ref{tab:execution_time} presents the average execution time per sample for RISC-V-based training on the CogDist and CarEvaluation datasets, comparing the performance of the GAP-8 V1 platform when running the same ML model using {\textit{float32}} and {\textit{int8}} representations.
In the CogDist dataset, the execution time is reduced $1.89\times$ (from 17 ms to 9 ms) when switching from {\textit{float32}} to {\textit{int8}} training. For the CarEvaluation dataset, execution time is reduced $2.2\times$ (from 11 ms to 5 ms).
These reductions demonstrate the significant computational advantages of {\textit{int8}} quantization, particularly on resource-constrained platforms such as the GAP-8, which does not include a hardware FPU, in contrast to the ARM-based MCU.

\begin{table}[t]
\centering
\caption{Average execution time per sample for RISC-V training.}
\label{tab:execution_time}
\begin{tabular}{l@{\hskip 1cm}l@{\hskip 1cm}c}
\toprule
\textbf{Dataset} & \textbf{L-SGD Configuration} & \textbf{Latency} \\
\midrule
\multirow{2}{*}{CogDist} 
  & (float32) RISC-V opt\_l-sgd         & 17 ms \\
  & (int8) \space \space \space \space RISC-V opt\_l-sgd            & 9 ms  \\
\midrule
\multirow{2}{*}{CarEvaluation} 
  & (float32) RISC-V opt\_l-sgd         & 11 ms \\
  & (int8) \space \space \space \space RISC-V opt\_l-sgd            & 5 ms  \\
\bottomrule
\end{tabular}
\end{table}

\subsection{Discussion}

This work demonstrates the feasibility of deploying on-device training of ML models on RISC-V MCUs, achieving stable convergence and competitive performance without significant degradation, even in the absence of hardware features such as FPUs. 
Through a tailored extension of the L-SGD optimizer, {\textit{float32} training was successfully executed directly on resource-constrained RISC-V platforms.
In addition to the full-precision implementation, a quantized version of L-SGD was developed and deployed, significantly reducing memory usage and training time. 
By leveraging {\textit{int8}} arithmetic for most operations while retaining floating-point representation for critical computations, the quantized implementation achieves nearly $4\times$ memory reduction and a $2.2\times$ speedup over the float baseline. These improvements establish quantized training as a practical solution for low-power systems. 
Nonetheless, this quantized solution introduces new constraints. Due to the risk of numerical saturation during training, the quantized pipeline is not currently suitable for training from scratch. Instead, it is best applied in FL scenarios, where the local device fine-tunes a pre-trained global model. This allows quantization to be leveraged during the personalization phase without compromising numerical stability.

\section{Conclusion}

This work demonstrates that training ML models on RISC-V MCUs, despite the absence of a hardware FPU, is both feasible and effective. The results confirm that RISC-V supports not only inference but also full-precision and quantized training of DNNs under tight resource constraints. The proposed quantized training approach achieves nearly $4\times$ memory reduction and a $2.2\times$ speedup in training time while maintaining all performance metrics. Overall, these findings highlight the viability of RISC-V as a platform for on-device learning with fine-tuning of pre-trained models in edge environments. 

Future work includes exploring more robust quantization techniques and broadening empirical evaluation beyond CogDist and CarEvaluation. Moreover, Decentor-V will be extended to support neural network architectures beyond DNNs, such as convolutional neural networks (CNNs), and to evaluate it on a wider range of RISC-V platforms, including those with hardware FPUs. These directions aim to further improve the applicability and efficiency of on-device training for next-generation RISC-V-based IoT systems.

%
%
%
%

\end{document}